\documentclass[12pt]{article}

\usepackage{newtxtext,newtxmath}

\usepackage{graphicx}
\usepackage{multirow}

\usepackage[letterpaper,margin=1in]{geometry}

\renewenvironment{abstract}
{\quotation}
{\endquotation}

\date{}

\makeatletter
\renewcommand{\fnum@figure}{\textbf{Figure \thefigure}}
\renewcommand{\fnum@table}{\textbf{Table \thetable}}
\makeatother

\usepackage{scicite}

\usepackage{url}

\usepackage{amsmath}
\usepackage{amsthm}
\usepackage{booktabs}
\usepackage[ruled,linesnumbered]{algorithm2e}

\theoremstyle{plain}

\theoremstyle{definition}

\theoremstyle{remark}

\usepackage{lineno}

\def\scititle{Probabilistic Energy Modeling Unifies Versatile Swarm Control}

\title{\bfseries \boldmath \scititle}

\author{
  Xiang Zhou$^{1}$,
  Xiangyu Zhu$^{1}$,
  Xintong Wang$^{1}$,
  Qingyuan Yang$^{1}$,
  \and
  Rui Huang$^{1}$,
  Feng Xue$^{1}$,
  Lin Li$^{1}$,
  Qingrui Zhang$^{1\ast}$\and
  \small$^{1}$School of Aeronautics and Astronautics, Sun Yat-sen University, Shenzhen 518107, China.\and
  \small$^\ast$Corresponding author. Email: zhangqr9@mail.sysu.edu.cn\and
}

\begin{document}

\maketitle

\begin{abstract} \bfseries \boldmath
Robotic swarms must coordinate across diverse missions, yet existing controllers often require redesign, retuning, or retraining when objectives change. Here we show that probabilistic energy modeling provides a unifying representation for versatile swarm control, in which diverse missions emerge from compositions of shared interaction principles rather than task-specific control formulations. By exploiting the correspondence between Gibbs random field energy minimization and structured reward maximization, we encode fundamental swarm behaviors as reusable energy terms within a unified probabilistic objective. Decentralized policies trained within this representation are composed into a single controller through lightweight policy gating, without retraining. An interpretable action-attention mechanism captures non-uniform local interactions, while dimensionless normalization supports transfer across swarm scales and platforms. Demonstrated on aerial swarms, the unified policy generalizes zero-shot across six mission classes, 10–100 agents, two hardware platforms, and indoor cluttered and GPS-denied woodland environments. Large-scale simulations achieve 0.96–1.00 task success rates with 1–10 ms per-agent inference latency, while physical deployments confirm zero-shot transfer to real platforms and unstructured environments. These results establish probabilistic energy modeling as a scalable and transferable framework for decentralized swarm control.

\end{abstract}

\noindent

\section*{Introduction}

Biological swarms coordinate collective motion while switching fluidly among flocking, foraging, and navigation through local interactions \cite{couzin_effective_2005, herbert-read_inferring_2011, rosenthal_revealing_2015, berdahl_emergent_2013, ballerini_interaction_2008}. This functional flexibility has long motivated robotic swarms, which combine multiple autonomous vehicles for exploration and monitoring and offer potential advantages in robustness, scalability, and efficiency over single-robot systems \cite{brambilla_swarm_2013, chung_survey_2018, mcguire_minimal_2019, zhou_swarm_2022}. A substantial gap nevertheless remains between the heterogeneous demands of multi-phase missions and current swarm systems, which are largely organized around task-specific designs \cite{brambilla_swarm_2013, chung_survey_2018}. Controllers are typically developed and evaluated around a single coordination objective, such as optimized flocking \cite{vasarhelyi_optimized_2018, zhu_heuristic_2025} or formation control \cite{honig_trajectory_2018, rezeck_chemistryinspired_2022}, and their reuse across heterogeneous missions is often limited. Onboard resource constraints, together with uncertain sensing and communication, further complicate coordinated motion across environments \cite{chung_survey_2018, soria_predictive_2021, tordesillas_mader_2022}. The central challenge is therefore not to improve any single-task controller, but to identify a shared mathematical representation in which diverse mission objectives can be expressed and reused, and to train a reusable decentralized policy that retains real-time execution.

Existing swarm control approaches face a persistent trade-off among coordination performance, computational demand, and cross-task reuse. Rule-based methods, including Reynolds-style interaction primitives, velocity vector fields, and artificial potential fields, provide decentralized scalability and low computational cost \cite{vasarhelyi_optimized_2018, guo_collisionfree_2023, reynolds_flocks_1987, mcguire_viscoelastic_2022}, but their heuristic structure usually requires scenario-specific tuning and can degrade when operating conditions depart from design assumptions \cite{guo_collisionfree_2023, soria_predictive_2021, zhou_swarm_2022}. Optimization-based frameworks, particularly Model Predictive Control, explicitly incorporate vehicle dynamics, safety constraints, and task objectives and achieve strong performance in specialized settings \cite{soria_predictive_2021, soria_distributed_2022, honig_trajectory_2018, tordesillas_mader_2022, zhu_heuristic_2025, yu_grfbased_2024, arul_swarmcco_2021}. However, online optimization and neighbor-trajectory exchange must fit limited onboard resources and real-time execution requirements \cite{soria_distributed_2022, tordesillas_mader_2022, senbaslar_dream_2025a}. Learning-based swarm control, including multi-agent reinforcement learning (MARL), shifts much of the computational burden offline and enables high-speed neural inference during deployment \cite{lowe_multiagent_2017, tolstaya_learning_2020, khan_graph_2020, batra_decentralized_2022, shi_neuralswarm2_2022, wang_learning_2020}. While learning-based flocking control and its performance trade-offs have been established in prior work \cite{zhang_learning_2025}, learned policies are typically optimized for individual tasks: extending them to new objectives often requires redesigning rewards, retraining policies, or deploying separate controllers. Thus, despite advances in efficient decentralized control and policy learning, a common representation that supports systematic reuse across heterogeneous swarm objectives remains less explored. The present work moves beyond single-task learned flocking toward a unified representation that supports multiple mission classes, modular policy integration, cross-platform deployment, and interpretable interaction analysis. This cross-task reuse bottleneck sits within a broader trend toward modular and compositional policy architectures in artificial intelligence \cite{shazeer_moe_2017, lee_multigame_2022, brohan_rt1_2023, driess_palme_2023}. For swarm systems, it must be reconciled with structural coherence during real-world deployment and sim-to-real transfer under environmental uncertainty \cite{shi_neuralswarm2_2022, soria_predictive_2021, zhou_swarm_2022}, while retaining a simple interface for high-level human or task-level guidance \cite{kolling_human_2016}.

Here we show that probabilistic energy modeling provides a unifying representation for this versatility. We build on Gibbs Random Fields (GRFs), in which swarm coordination objectives are expressed as a common probabilistic energy composed of reusable unary terms for robot-to-environment interactions and pairwise terms for robot-to-robot interactions. Cohesion, collision avoidance, velocity alignment, formation regulation, and task-specific guidance are encoded as energy terms within one shared objective rather than as independent rewards designed for each task. We leverage the direct correspondence between GRF energy minimization and reward maximization: minimizing the Gibbs energy corresponds to maximizing a structured reward, so the same energy formulation yields consistent reward functions across missions. These energy-based rewards train decentralized specialist policies with MARL, and an action-attention mechanism enables interpretable, non-uniform weighting of neighboring agents while preserving fully decentralized execution and inference independent of swarm size. Dimensionless observation normalization provides a scale-consistent interface for high-level commands. The result is a single integrated controller, from which distinct mission capabilities arise by composing shared coordination energy terms with task-specific guidance potentials, and new capabilities can be incorporated without modifying previously learned policies. Although demonstrated on aerial platforms, the energy-based formulation is not inherently tied to a specific robotic platform. By unifying diverse objectives within a single probabilistic energy framework and demonstrating zero-shot transfer across tasks, scales, and platforms, this work establishes a representation layer upon which large-scale pre-trained swarm policies—analogous to foundation models in language and vision—could be constructed in the future.

We validate the framework through large-scale simulation benchmarks and physical deployments spanning four axes of generalization: mission class, swarm scale, hardware platform, and environment. In simulation, the unified policy attains task-execution success rates of $0.96$--$1.00$ across traversal settings ranging from sparse $10$-agent, $10$-obstacle scenarios to congested $100$-agent, $50$-obstacle configurations, while per-agent inference remains $O(1)$ in swarm size at $1$--$10$ ms. Physical deployments on $29$-g Crazyflie micro-quadrotors and custom-built outdoor quadrotors evaluate traversal, high-speed racing, persistent patrol, pattern formation, distributed exploration, and an integrated multi-phase outdoor workflow—without retraining or fine-tuning. Experiments span indoor passages, obstacle-dense arenas, and GPS-denied woodlands, with the same policy transferring directly across platforms and environments. Collectively, these results demonstrate zero-shot generalization across six mission classes, swarm scales from 10 to 100 agents, two hardware platforms, and indoor and GPS-denied outdoor environments, supporting reusable decentralized swarm coordination across diverse operating conditions.

\section*{Results}

\subsection*{Method overview}

\begin{figure}[tbph]
\centering
\includegraphics[width=1.0\textwidth]{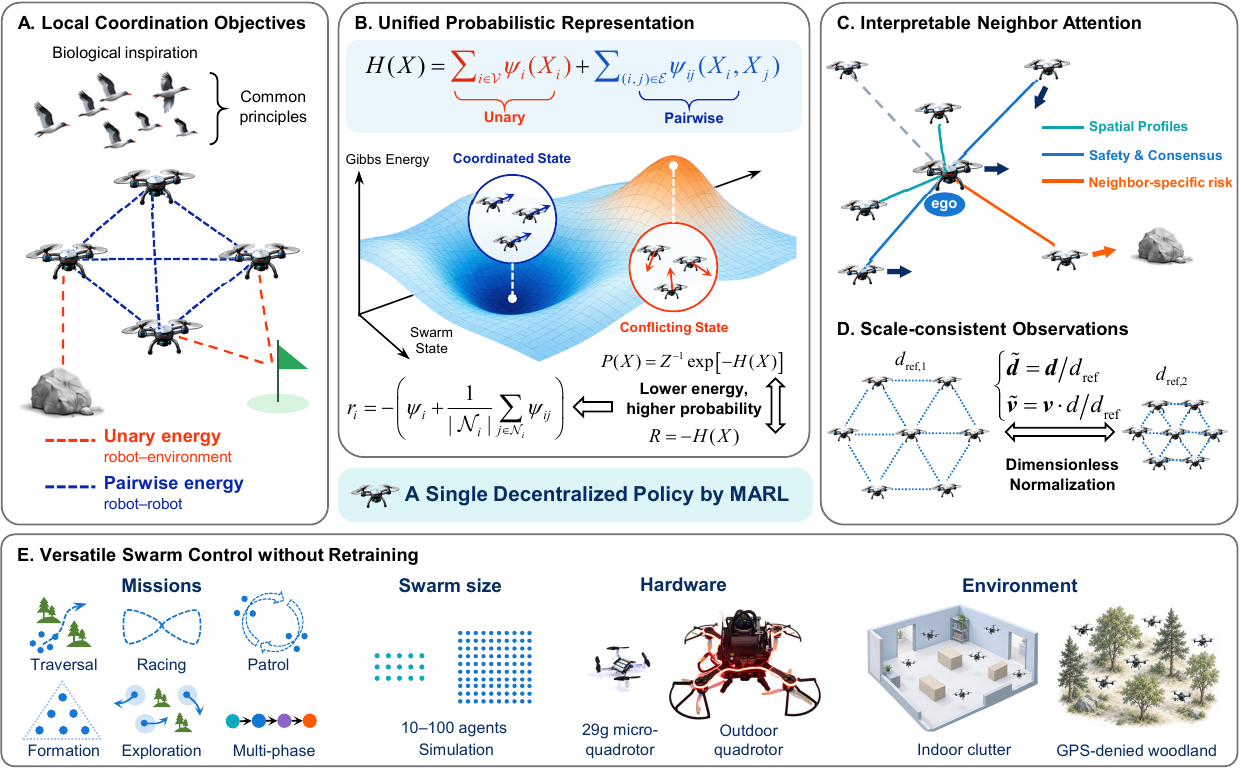}
\caption{\textbf{Probabilistic energy modeling for versatile swarm control.}
\textbf{(A)} Inspired by biological swarms, diverse non-contact swarm tasks are described through several reusable local coordination objectives, encoded as unary and pairwise energy terms.
\textbf{(B)} A shared Gibbs energy defines swarm-state probabilities $P(X)\propto\exp[-H(X)]$ and the reward $R=-H(X)$ for multi-agent reinforcement learning.
\textbf{(C)} Learned neighbor attention varies with spatial organization, safety and motion consensus, and neighbor-specific risk.
\textbf{(D)} Reference-distance normalization provides scale-consistent observations for policy reuse.
\textbf{(E)} The learned controller is deployed across tasks, swarm sizes, UAV platforms, and environments without retraining.}
\label{fig:fig1_framework}
\end{figure}

To unify disparate swarm-control objectives, we model the collective system through GRFs \cite{koller_probabilistic_2009}, an undirected graphical model in which each agent is a random variable and local interactions are encoded through unary and pairwise energy terms (Fig.~\ref{fig:fig1_framework}). The joint distribution takes a Gibbs form $P(X) \propto \exp{\left[-H(X)\right]}$, where the total energy $H(X)$ decomposes into unary terms $\psi_i$ for agent-to-environment interactions, such as goal tracking and obstacle avoidance, and pairwise terms $\psi_{ij}$ for agent-to-agent interactions, such as cohesion, velocity alignment, and collision avoidance. This decomposition establishes a direct correspondence with potential-based reward shaping: minimizing the Gibbs energy is equivalent to maximizing the structured reward $R=-H(X)$. Cohesion, collision avoidance, velocity alignment, pattern regulation, and task-specific guidance can therefore be represented as reusable energy terms within a shared probabilistic objective, with different missions obtained by reweighting these terms rather than redesigning task-specific rewards.

These energy-based rewards train decentralized specialist policies through multi-agent reinforcement learning under a Dec-POMDP formulation with Decentralized Training with Decentralized Execution (DTDE). Each agent uses local observations of its own state, neighboring states within a sensing radius, and sector-based obstacle distances. To reconcile the global flock-energy objective with local observations, we derive a decoupled pairwise energy $\hat{H}_p$ that approximates the global objective using only local neighbor interactions (Proposition~\ref{prop:edge_weight}). A neighbor action-attention mechanism then selectively aggregates neighboring behavioral information according to the current interaction context, capturing non-uniform interactions while preserving fully decentralized execution and $O(1)$ per-agent inference complexity in swarm size. An integration module implemented as a lightweight gating network combines the independently trained and frozen specialist policies conditioned on local observations and a high-level task command, enabling a single integrated controller while preserving policy specialization and avoiding retraining.

For cross-scale and cross-platform transfer, dimensionless observation normalization maps spatial and temporal quantities to a reference inter-agent distance $d_{\mathrm{ref}}$ and a reference timescale, providing a scale-consistent policy interface. The unified policy serves as a reusable low-level controller for decentralized execution: primitive behaviors are realized through combinations of specialist policies, while complex missions such as persistent patrol and distributed exploration operate within a hierarchical framework in which existing planners provide navigation goals or region assignments and the learned policy performs coordinated local motion. This separation allows new capabilities to be incorporated through additional specialist policies without modifying previously learned policies, while maintaining decentralized real-time execution across heterogeneous operating conditions.

\subsection*{Elementary behaviors in simple scenarios}

\begin{figure}[tbph]
  \centering
  \includegraphics[width=0.9\textwidth]{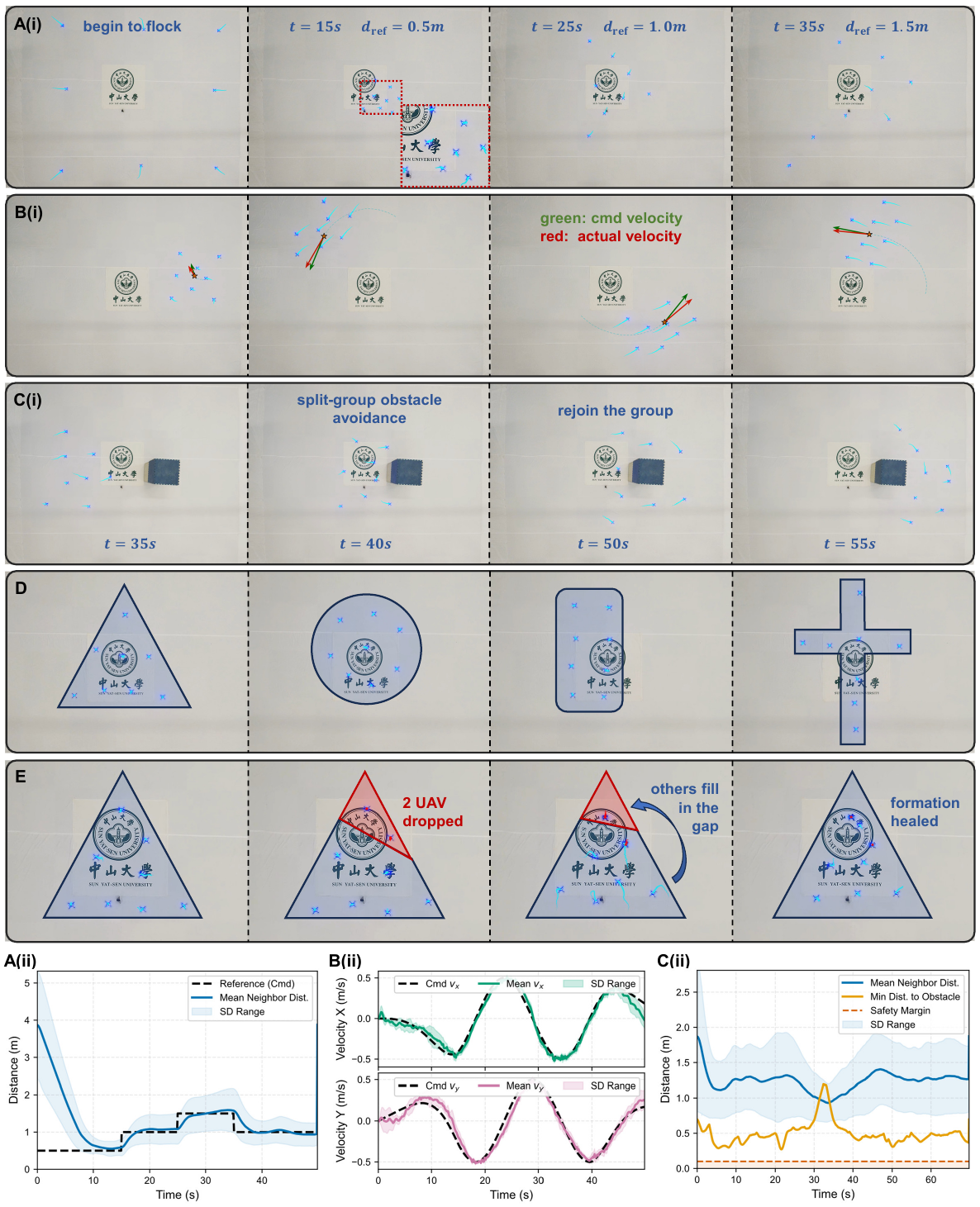}
  \caption{\textbf{Elementary behaviors in simple scenarios.}
    (\textbf{A}) Cohesive aggregation and inter-agent spacing control. (i) Swarm aggregation under spacing commands. (ii) Neighbor distance response.
    (\textbf{B}) Velocity tracking. (i) Swarm motion under time-varying velocity commands. (ii) Commanded and measured centroid velocities.
    (\textbf{C}) Collision avoidance. (i) Swarm deformation to bypass an obstacle. (ii) Temporal evolution of inter-agent and obstacle-clearance distances.
    (\textbf{D}) Pattern formation over diverse geometries.
    (\textbf{E}) Reconfiguration after agent dropout in a pattern-formation test.}
  \label{fig:basic_policy_tests}
\end{figure}

We begin with the elementary behaviors that underpin swarm missions—aggregation, velocity tracking, collision avoidance, geometric deployment, and reconfiguration after agent dropout—and verify that the unified policy reproduces them in simple closed-loop scenarios (Fig.~\ref{fig:basic_policy_tests}). The policy drives a sparsely distributed group into a cohesive swarm and tracks successive step changes in the reference inter-neighbor distance $d_{\mathrm{ref}}$ with minimal latency, while the standard deviation of neighbor spacing remains below $0.5$ m throughout (Fig.~\ref{fig:basic_policy_tests}A). It also follows a circular velocity command with a radius of $1.7$ m through acceleration (0–15 s), cruise (15–40 s), and deceleration (40–50 s) phases, with the swarm-centroid velocity closely tracking the multi-axis command with minimal phase lag (Fig.~\ref{fig:basic_policy_tests}B). Under an approaching obstacle on its right ($\sim$32 s; Fig.~\ref{fig:basic_policy_tests}C), the swarm splits into two subgroups and deforms its topology to pass, while its mean neighbor spacing increases only slightly, preserving global connectivity without compromising safety; after the obstacle is cleared ($\sim$50 s), the subgroups rejoin. The minimum UAV–obstacle clearance remains above $0.25$ m, well beyond the $0.1$ m safety radius. The same policy assembles non-convex geometric patterns (Fig.~\ref{fig:basic_policy_tests}D). After a subset of agents is removed, the policy redistributes the remaining agents to fill the vacated regions and restore the target morphology (Fig.~\ref{fig:basic_policy_tests}E). No behavior requires a hand-tuned controller or task-specific retraining; these elementary responses emerge from the unified architecture and provide the coordination primitives for the multi-mission evaluations below.

\subsection*{Versatile swarm control across missions}

\begin{figure}
  \centering
  \includegraphics[width=0.95\textwidth]{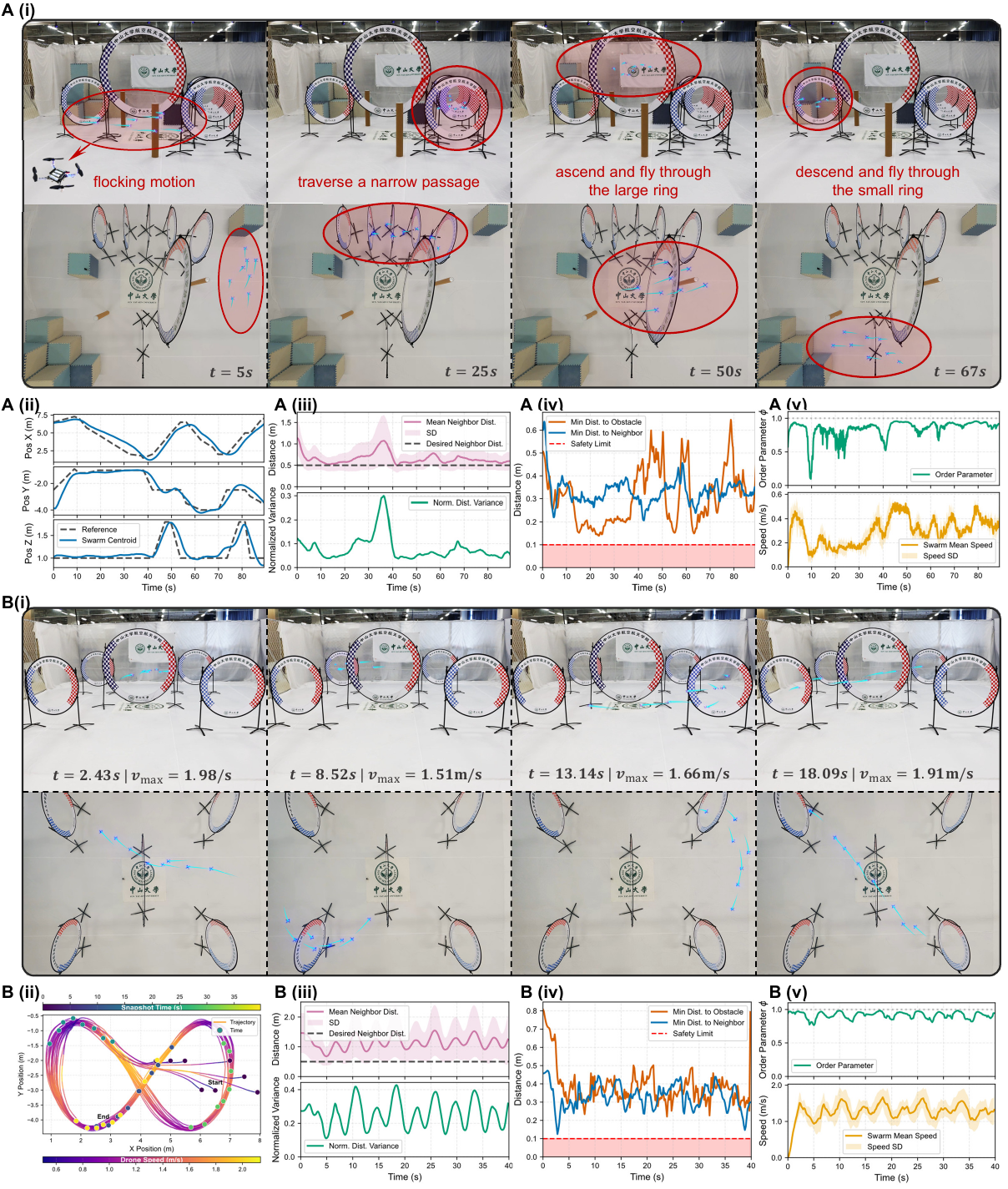}
\caption{\textbf{Traversal and racing performance in cluttered indoor environments.}
    (\textbf{A}) Traversal. (i) Swarm traversing through gates. (ii) Centroid tracking result.
    (\textbf{B}) Racing. (i) Swarm racing through sequential gates. (ii) Racing trajectories color-coded by instantaneous speed.
    A(iii) and B(iii) Mean neighbor distance and normalized variance to quantify cohesion and spatial uniformity.
    A(iv) and B(iv) Minimum neighbor and obstacle clearances.
    A(v) and B(v) Order parameter and mean speed.}
  \label{fig:fig3_traversal_racing}
\end{figure}

Building on these elementary behaviors, we next ask whether the unified policy can support qualitatively different missions under varying demands on motion, spatial organization, sensing, and resilience. We consider traversal and racing in cluttered environments, persistent patrol and pattern formation, and distributed exploration under agent loss and changing swarm scale. Across these tasks, the same learned policy adapts its collective configuration and local interactions to task demands without task-specific retraining.

\textbf{Traversal and racing.}
Traversal and racing probe two distinct regimes of coordinated motion: constrained navigation and high-speed trajectory tracking (Fig.~\ref{fig:fig3_traversal_racing}). In traversal, eight Crazyflie quadrotors pass through four sequential phases—formation assembly, compression through a narrow gate corridor, vertical ascent over a central gate, and coupled yaw-and-descent through the terminal gate (Fig.~\ref{fig:fig3_traversal_racing}A(i)). The swarm deforms its topology in narrow passages while maintaining collision-free motion and centroid tracking (Fig.~\ref{fig:fig3_traversal_racing}A(ii)). Except for corridor-induced elongation between 20--40\,s, the mean inter-agent distance remains close to the 0.5\,m equilibrium, with normalized variance below 0.1 (Fig.~\ref{fig:fig3_traversal_racing}A(iii)). Minimum agent--agent and agent--obstacle clearances remain above the $0.05$ m collision threshold (Fig.~\ref{fig:fig3_traversal_racing}A(iv)). The order parameter $\phi$ decreases during dense local avoidance and recovers toward 1 as the swarm enters the sparse region (Fig.~\ref{fig:fig3_traversal_racing}A(v)).

Racing imposes a different coordination regime. The swarm passes five circular gates at high speed (Fig.~\ref{fig:fig3_traversal_racing}B(i)), reaching $2$~m/s on straight segments and decreasing to approximately $0.8$~m/s during sharp turns (Fig.~\ref{fig:fig3_traversal_racing}B(ii)). At these higher speeds, the swarm adopts an elongated ribbon topology, increasing mean neighbor distance and its variance relative to traversal (Fig.~\ref{fig:fig3_traversal_racing}B(iii)). Although inter-agent clearance approaches the $0.05$ m collision threshold, obstacle clearance remains positive throughout the $40$ s trial (Fig.~\ref{fig:fig3_traversal_racing}B(iv)). Speed and $\phi$ vary in phase around the gates: cornering reduces both, whereas straight-line acceleration restores higher speed and alignment (Fig.~\ref{fig:fig3_traversal_racing}B(v)). Thus, the policy preserves the underlying coordination constraints while allowing the collective geometry to adapt from compact navigation to elongated high-speed motion.

\begin{figure}
  \centering
  \includegraphics[width=1.0\textwidth]{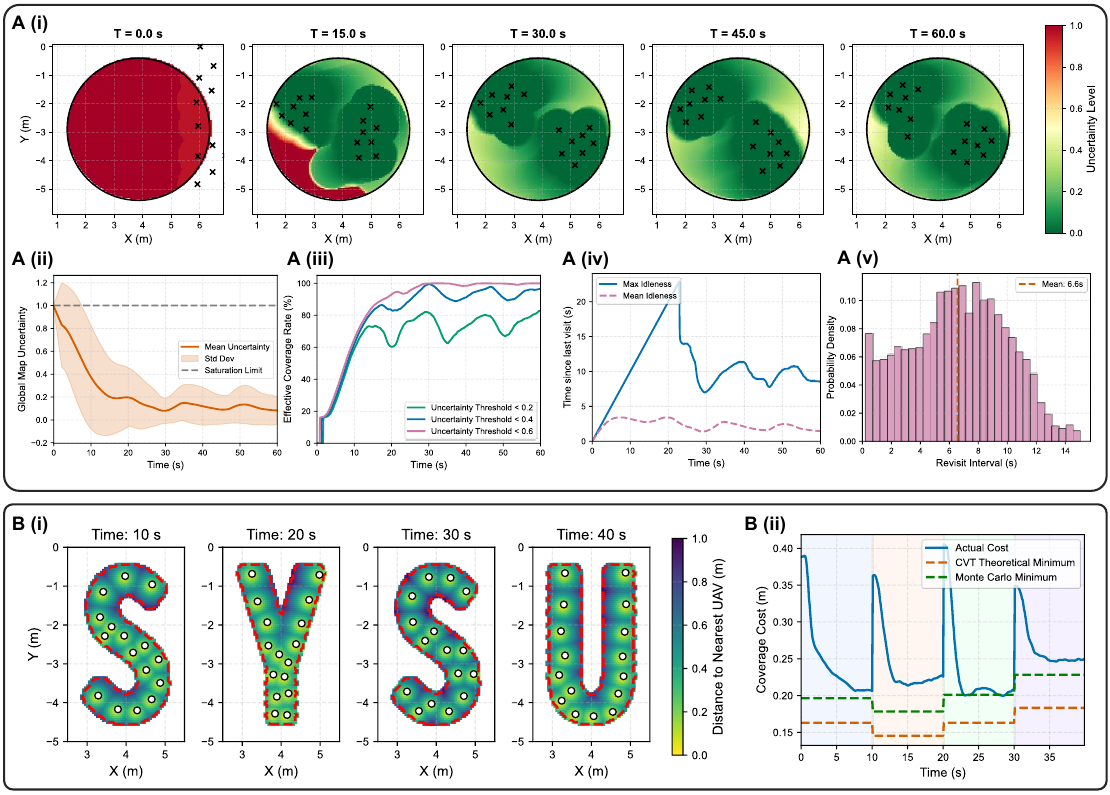}
    \caption{\textbf{Persistent patrol and pattern formation.} (\textbf{A}) Persistent patrolling with information decay. (i) Spatio-temporal uncertainty maps during persistent patrolling by two sub-swarms. (ii) Global uncertainty statistics, showing rapid decay to a low steady state. (iii) Effective coverage rates under different uncertainty thresholds. (iv) Maximum and mean idleness, characterizing revisit timeliness. (v) Revisit-interval distributions across the monitored region. (\textbf{B}) Pattern formation over non-convex geometries. (i) Sequential configurations for the ``S--Y--S--U'' target shapes, with distance-to-nearest-agent fields indicating spatial uniformity. (ii) Coverage cost $J(t)$ during shape transitions, showing convergence toward near-optimal baselines, with a larger residual error for the more challenging ``Y'' geometry.}
  \label{fig:fig4_patrol_formation}
\end{figure}

\textbf{Persistent patrol and pattern formation.} We next examine behaviors dominated by spatial allocation rather than forward motion (Fig.~\ref{fig:fig4_patrol_formation}). In persistent patrol within a circular region of radius $r=2.5$\,m, uncertainty increases at $0.05$ per second wherever fewer than two UAVs simultaneously provide coverage, each with a sensing radius of $0.75$\,m (1.5 times the reference spacing). Starting from maximum entropy, the swarm autonomously partitions into subgroups and drives global uncertainty to a steady mean of $0.1$--$0.2$ (Fig.~\ref{fig:fig4_patrol_formation}A(i--ii)). Effective coverage remains near $100\%$ for thresholds $0.4$ and $0.6$ and approximately $80\%$ for the $0.2$ threshold (Fig.~\ref{fig:fig4_patrol_formation}A(iii)). After full deployment at $T\approx23$\,s, maximum idleness remains within $7$--$11$ s and mean idleness within $2$--$3$ s, with a mean revisit interval of $6.6$ s (Fig.~\ref{fig:fig4_patrol_formation}A(iv--v)).

For pattern formation, $N=16$ UAVs reconfigure across non-convex ``S--Y--S--U'' shapes within $40$ s under collision- and deadlock-free transitions (Fig.~\ref{fig:fig4_patrol_formation}B(i)). Because the Centroidal Voronoi Tessellation optimum assumes an infinite continuous plane and does not account for finite-size effects in a small non-convex swarm, we use Monte Carlo sampling as a finite-domain baseline. Following commands at $T=0,10,20,30$ s, the coverage cost $J(t)$ converges toward this baseline (Fig.~\ref{fig:fig4_patrol_formation}B(ii)): ``S'' and ``U'' approach the sampled minimum, whereas ``Y'' retains a small steady-state error, likely reflecting the difficulty of resolving its intersecting junctions through local interactions. This behavior is consistent with the underlying Gibbs energy formulation, which favors low-energy configurations under the prescribed spatial objective.

\begin{figure}
  \centering
  \includegraphics[width=1.0\textwidth]{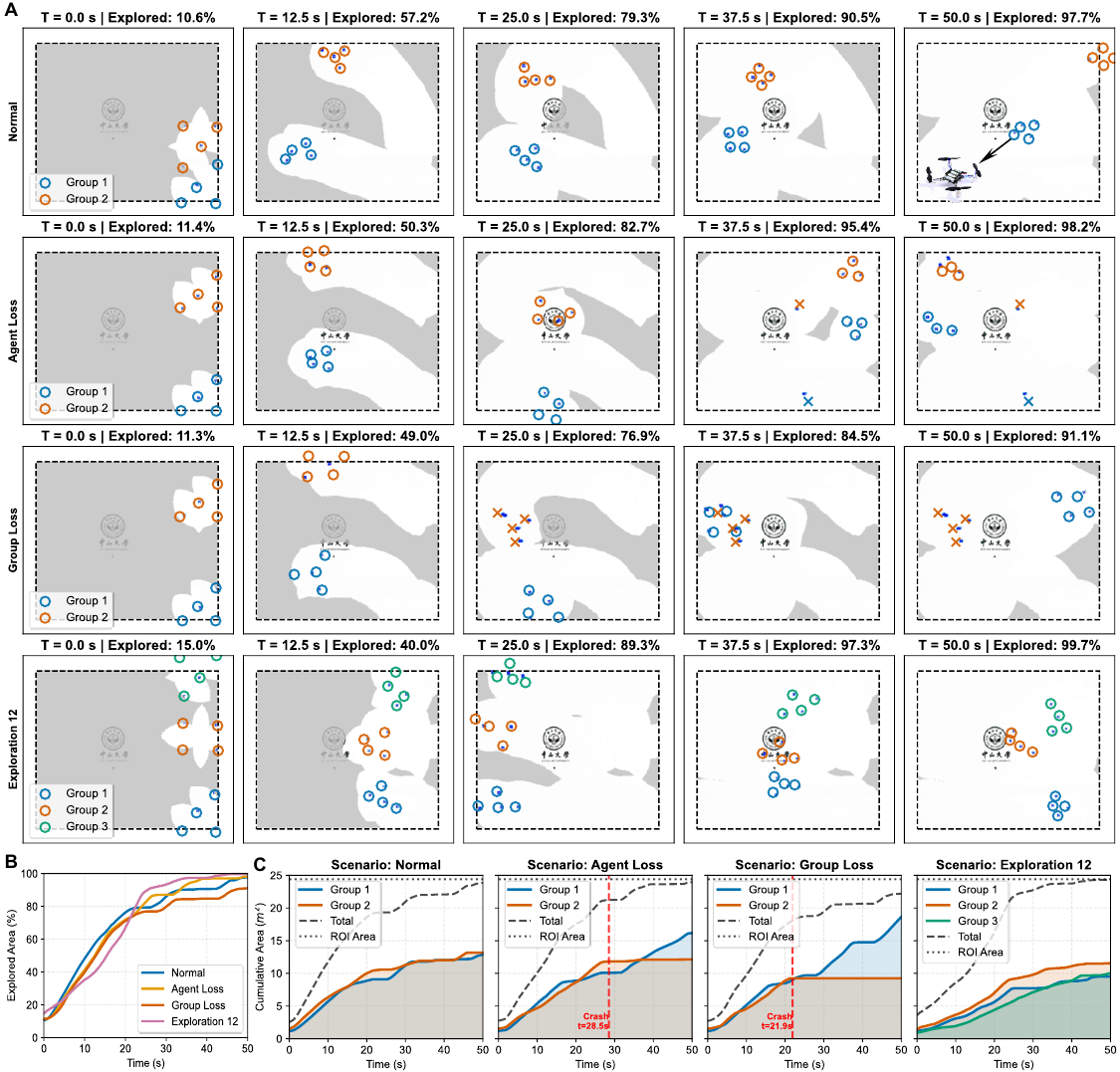}
    \caption{\textbf{Distributed exploration under agent loss and varying swarm sizes.} (\textbf{A}) Exploration snapshots for four scenarios: (i) normal operation with 8 UAVs divided into two subgroups; (ii) individual agent loss in each subgroup; (iii) complete loss of one subgroup; and (iv) exploration with 12 UAVs divided into three subgroups. (\textbf{B}) Total explored-area ratio over time for all scenarios. (\textbf{C}) Cumulative explored area contributed by each subgroup, showing the distribution of exploration contributions across scenarios.}
  \label{fig:fig5_exploration}
\end{figure}

\textbf{Distributed exploration under failures and scale variation.} Finally, we test whether the learned coordination layer remains effective under limited sensing and changing swarm composition (Fig.~\ref{fig:fig5_exploration}). A hierarchical planner supplies cooperative objectives while the coordination layer executes local motion. Each UAV has a $0.75$ m sensing radius, and a location counts as explored only when covered concurrently by at least two UAVs. With eight UAVs, the swarm splits into two subgroups and reaches $97.7\%$ coverage within $50$ s with comparable contributions from the two subgroups (Fig.~\ref{fig:fig5_exploration}A(i),B--C).

The same coordination layer adapts to agent and subgroup loss. When one UAV in each subgroup fails at $T=28.5$\,s, the survivors redistribute and reach $98.2\%$ coverage without an evident interruption in cumulative coverage (Fig.~\ref{fig:fig5_exploration}A(ii),B--C). When an entire subgroup drops out at $T=21.9$\,s, exploration slows but continues without deadlock, reaching 91.1\% coverage with the remaining subgroup (Fig.~\ref{fig:fig5_exploration}A(iii),B--C). Increasing to 12 UAVs in three subgroups reveals a density--congestion trade-off: denser initial deployment slows early exploration, but the additional sensing capacity ultimately yields $99.7\%$ coverage (Fig.~\ref{fig:fig5_exploration}A(iv),B).

Overall, these experiments show that the unified policy provides reusable coordination across distinct motion, spatial-allocation, and exploration regimes, with task-dependent collective configurations emerging without task-specific retraining, within the tested conditions.

\subsection*{Benchmarking and interpreting local interactions}

\begin{figure}[tbph]
  \centering
  \includegraphics[width=1.0\textwidth]{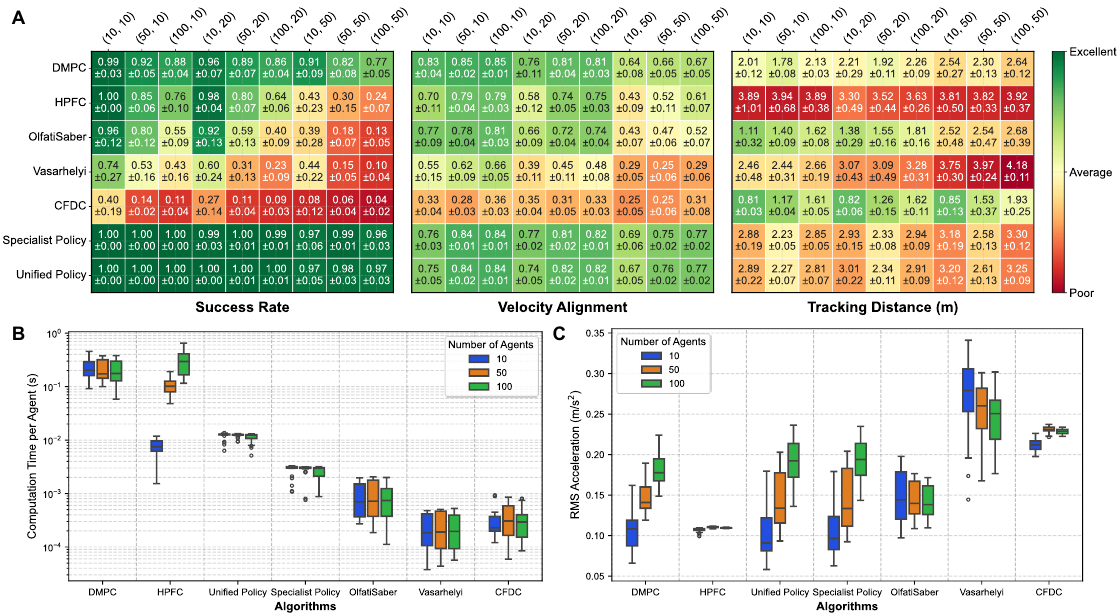}
  \caption{\textbf{Quantitative performance benchmarking.}
    (\textbf{A}) Success rate, velocity alignment, and tracking distance for swarm sizes from $10$ to $100$ agents and obstacle densities from $10$ to $50$, evaluated over ten random seeds. Cells show mean $\pm$ s.d. (\textbf{B}) Per-agent computation time on a logarithmic scale, showing approximately size-independent inference for the unified policy compared with increasing computation time for the baselines. (\textbf{C}) RMS acceleration as a measure of trajectory smoothness.}
  \label{fig:fig6_compare}
\end{figure}

We next benchmark the unified policy against established controllers and analyze the learned neighbor interactions. We compare optimization-based controllers (DMPC, HPFC) and rule-based approaches (Olfati--Saber, V\'{a}s\'{a}rhelyi, CFDC) across $10$--$100$ agents and $10$--$50$ obstacles
over ten random seeds (Fig.~\ref{fig:fig6_compare}).

\textbf{Quantitative benchmarking.}
The unified policy maintains success rates of approximately $0.96$--$1.00$ across the tested conditions, including the most congested settings (Fig.~\ref{fig:fig6_compare}A). Rule-based methods deteriorate with obstacle density: CFDC falls below $0.10$, consistent with conflicting repulsion creating local minima, and its small tracking distance reflects early collision-induced termination rather than improved tracking. DMPC and HPFC remain safe in sparse settings but degrade under congestion as their optimization problems become infeasible. The unified policy also sustains velocity alignment of $0.74$--$0.84$, indicating that coordinated deformation through narrow passages does not sacrifice motion coherence. The unified policy matches the corresponding specialist policy on every metric, with no measurable degradation from multi-task integration.

Per-agent inference stays within $1$--$10$\,ms across $10$--$100$ agents (Fig.~\ref{fig:fig6_compare}B), decoupling swarm cardinality from computational overhead at $\mathcal{O}(1)$ scaling; the integration module adds marginal cost. DMPC carries high baseline latency, and HPFC grows rapidly with swarm size. RMS acceleration stays below $0.22$~m/s$^2$ (Fig.~\ref{fig:fig6_compare}C), versus $>0.30$~m/s$^2$ for V\'{a}s\'{a}rhelyi; HPFC achieves low RMS but at the cost of computation and dense-setting failures. Together, these results show stable task performance, low-latency decentralized inference, and smooth collective motion across a broad range of densities.

\begin{figure}[htbp]
  \centering
  \includegraphics[width=1.0\textwidth]{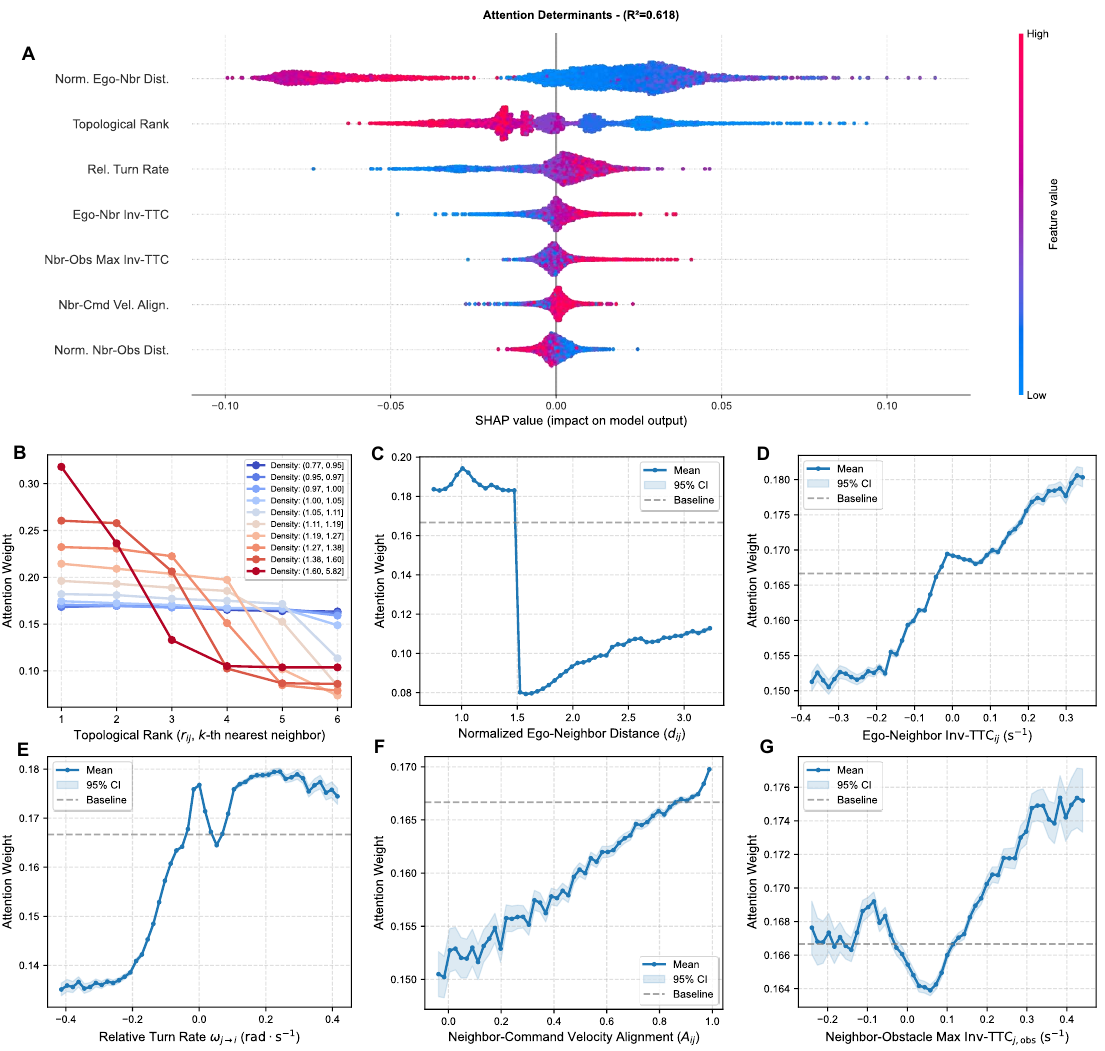}
  \caption{\textbf{Interpreting learned neighbor interactions.} (\textbf{A}) SHAP feature attribution of attention weights predicted by an XGBoost surrogate ($R^2=0.618$), ranked by feature importance. (\textbf{B}) Attention weight versus topological neighbor rank across local densities. (\textbf{C--G}) Attention responses to (C) normalized ego--neighbor distance, (D) ego--neighbor inverse time-to-collision, (E) relative turn rate, (F) neighbor--commanded-velocity alignment, and (G) neighbor--obstacle maximum inverse time-to-collision. Dashed lines indicate the uniform-attention level.}\label{fig:attention}
\end{figure}

\textbf{Interpreting learned interactions.}
We next examine how the policy allocates attention among its six nearest neighbors across traversal scenarios with varying swarm sizes and obstacle densities (Fig.~\ref{fig:attention}), using spatial, topological, kinematic, and collision-risk variables (Supplementary Methods). An XGBoost surrogate predicts attention weights with a held-out $R^2=0.618$, so SHAP values characterize the surrogate rather than the neural policy itself (Fig.~\ref{fig:attention}A). Ego--neighbor distance and topological rank dominate; relative turn rate toward the ego agent and ego--neighbor inverse time-to-collision (Inv-TTC) increase attention, whereas static neighbor--obstacle proximity is weak. Attention thus encodes both spatial organization and kinematic context, not distance alone. In sparse states, attention decays with topological rank, favoring the first three neighbors; near the desired density it approaches uniform weighting ($\alpha_{ij}\approx 1/6$), qualitatively consistent with topological interactions in starling flocks \cite{ballerini_interaction_2008}, though the six-neighbor limit is architectural rather than learned. Attention peaks at short range and near the reference spacing $d_{\mathrm{ref}}$ (Fig.~\ref{fig:attention}B,C). As with inferred interaction forces in fish schools \cite{katz_inferring_2011}, these weights encode information allocation rather than physical force.

Dynamic attention rises with ego--neighbor Inv-TTC and the neighbor's turn rate toward the ego agent (Fig.~\ref{fig:attention}D,E), qualitatively resembling speed- and heading-dependent responses in fish schools \cite{herbert-read_inferring_2011}, and increases with neighbor--command velocity alignment (Fig.~\ref{fig:attention}F), consistent with reinforcing local motion consensus rather than selecting informed leaders. Finally, attention rises toward neighbors facing high obstacle-collision risk even when the ego agent itself is safe (Fig.~\ref{fig:attention}G), suggesting that local coordination accounts for a neighbor's environmental context. This offers a qualitative parallel to collective environmental sensing in animal groups \cite{berdahl_emergent_2013, rosenthal_revealing_2015}, without implying identical mechanisms. Together, these profiles reveal interpretable associations between neighbor weighting and cohesion, safety, motion consensus, and environmental risk, without establishing causal effects on task performance or anticipation.

We next test whether the unified policy transfers to larger platforms and unstructured outdoor environments without retraining.

\subsection*{Zero-shot transfer across platforms and environments}

\begin{figure}
  \centering
  \includegraphics[width=1.0\textwidth]{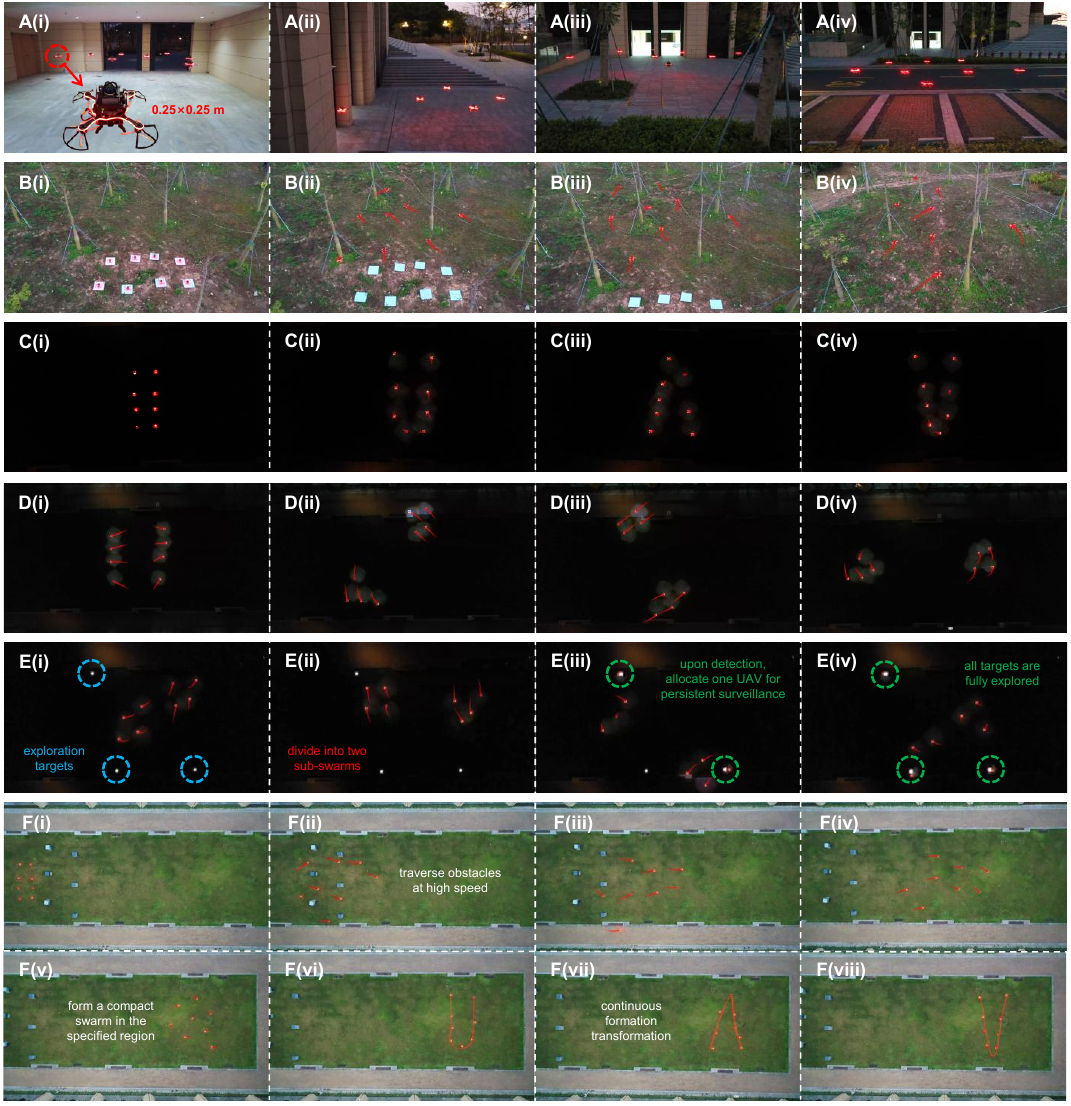}
  \caption{\textbf{Zero-shot field generalization.} (\textbf{A}) Cross-environment transition from an underground parking garage to an open outdoor area. (\textbf{B}) Traversal through dense, unstructured woodland. (\textbf{C--D}) Large-scale formation reconfiguration and subgroup cooperative patrol. (\textbf{E}) Decentralized exploration with persistent target monitoring in an unmapped area. (\textbf{F}) Integrated mission combining traversal, regrouping, and formation transformation.}\label{fig:fig8_outdoor}
\end{figure}

The generality of the learned coordination policy was further tested through zero-shot deployment beyond the training distribution.  Without fine-tuning or retraining, the same policy was deployed on custom-built quadrotors across indoor, open outdoor, and GPS-denied woodland environments (Fig.~\ref{fig:fig8_outdoor}), enabled by the dimensionless observation representation that provides a common input across physical scales.

Across indoor, open outdoor, and dense woodland environments, the swarm maintains coherent and collision-free motion despite changes in geometry, sensing conditions, and aerodynamic disturbances (Fig.~\ref{fig:fig8_outdoor}A,B). The policy also transfers from micro-quadrotors to larger custom-built platforms, which perform formation reconfiguration and subgroup cooperative patrol (Fig.~\ref{fig:fig8_outdoor}C,D). The same policy further adapts to changing mission conditions without modification. In an unmapped environment, eight UAVs divide into two subgroups for exploration; after detecting a target, one UAV maintains surveillance while the others continue searching (Fig.~\ref{fig:fig8_outdoor}E). Finally, the swarm combines traversal, regrouping, and formation transformation within a single mission (Fig.~\ref{fig:fig8_outdoor}F).

Together, these field experiments demonstrate that the unified policy transfers zero-shot across platforms, environments, and mission phases without task-specific retraining.

\section*{Discussion}

This work shows that probabilistic energy modeling can unify versatile swarm control. By encoding cohesion, collision avoidance, velocity alignment, formation regulation, and task guidance as structured terms in a single Gibbs energy, diverse coordination objectives can be expressed within a common reward-shaping framework. A single decentralized policy, trained once, generalizes across six mission classes, 10--100 agents, two aerial platforms, and indoor cluttered and GPS-denied woodland environments without retraining. Dimensionless observations provide a scale-consistent interface across platforms, while modular specialist policies enable new coordination capabilities to be incorporated without redesigning the policy.

The attention analysis further links this energy formulation to learned local interactions. Neighbor weighting varies with topological rank, proximity, relative motion, and environmental collision risk, indicating context-dependent information allocation. These patterns qualitatively echo interaction structures reported in biological collectives, but do not imply a shared mechanism. Importantly, attention represents information allocation rather than physical force, and the surrogate analysis establishes associations rather than causality. Together, these results suggest a principled way to combine structured coordination priors with modular policy learning, separating reusable swarm behaviors from task-specific specialization.

Several limitations remain. The current system depends on local state estimation and inter-agent communication, and formal safety guarantees under severe sensing, communication, or localization degradation remain unestablished. Local observations can also produce residual errors near sharp or non-convex boundaries, while physical congestion limits achievable swarm density in confined environments. Real-robot experiments scale to 16 quadrotors; simulation shows $O(1)$ latency up to $100$ agents, but physical validation at $100$ agents remains future work. Moreover, attention attribution does not establish whether specific interaction patterns improve task performance; targeted interventions are needed to address this causal gap. Future work should investigate robust multimodal perception, continual specialist-policy integration, and heterogeneous multi-robot teams to assess the scalability of probabilistic energy modeling beyond aerial platforms.

\section*{Methods}

\subsection*{Probabilistic energy formulation}

We formulate swarm coordination as a Gibbs random field (GRF), which provides a common probabilistic representation for diverse collective behaviors. A GRF defines a joint distribution over random variables $\{X_i\}_{i\in\mathcal{V}}$ through an undirected graph $\mathcal{G}=(\mathcal{V},\mathcal{E})$:
\begin{equation}
  P(X)=\frac{1}{Z}\exp\{-H(X)\},\quad
  H(X)=\sum_{c\in\mathcal{C}}\psi_c(X_c) \label{gibbs_distribution}
\end{equation}
where $\mathcal{C}$ is the set of cliques, $\psi_c$ the clique energy, and $Z$ the normalization constant. Treating each robot as a random particle, we restrict the graph to singleton and edge cliques for
tractability, neglecting higher-order interactions. The total energy then decomposes into unary robot--environment and pairwise robot--robot terms:
\begin{equation}
  H(\mathbf{O},\mathbf{A})=
\sum_{i\in\mathcal{V}}\psi_i(\boldsymbol{o}_i,\boldsymbol{a}_i) +\sum_{(i,j)\in\mathcal{E}}\psi_{ij}(\boldsymbol{o}_i,\boldsymbol{o}_j).
  \label{eq:unary_pairwise_potential}
\end{equation}
Because lower Gibbs energy corresponds to higher probability, energy minimization maps to reward maximization. We define the local reward as
\begin{equation}
  r_i=-\left(\psi_i+\frac{1}{|\mathcal{N}_i|}\sum_{j\in\mathcal{N}_i}\psi_{ij}\right)
  \label{eq:local_reward}
\end{equation}
where $\psi_i$ captures task objectives (target tracking, velocity maintenance, static obstacle avoidance) and $\psi_{ij}$ captures local inter-agent constraints (collision avoidance, velocity alignment, cohesion). The complete reward components are listed in Table~\ref{tab:reward_components}.

\subsection*{Decentralized learning with attention}

Each robot $i$ follows second-order dynamics and uses a hybrid action space $\mathcal{A}_i=\mathcal{A}_{i,d}\cup\mathcal{A}_{i,c}$: $\mathcal{A}_{i,d}$ contains fixed-magnitude acceleration vectors, while $\mathcal{A}_{i,c}$ incorporates damping and velocity-tracking heuristics. Under decentralized execution, robot $i$ does not observe its neighbors' instantaneous actions. We therefore adopt a mean-field approximation, representing neighbor influence through marginal action distributions and previous neighbor actions $\boldsymbol{A}_{j,t-1}$ as proxies for latent intentions.

Social context is encoded through an attention mechanism over the local neighborhood (Fig.~\ref{fig:figS4_network_structure}). For each ego--neighbor pair $(i,j)$, the relative state is embedded as $\boldsymbol{e}_{ij}=g(\boldsymbol{s}_{ij})$ and an interaction feature as $\boldsymbol{m}_{ij}=h(\boldsymbol{s}_i,\boldsymbol{s}_{ij})$, with $\boldsymbol{A}_{j,t-1}$ incorporated when available. Query, key, and value are
\begin{equation}
  \boldsymbol{q}_{ij}=W_q\boldsymbol{e}_{ij},\quad
  \boldsymbol{k}_{ij}=W_k\boldsymbol{m}_{ij},\quad
  \boldsymbol{l}_{ij}=W_v\boldsymbol{e}_{ij}
\end{equation}
and the normalized attention weight is
\begin{equation} \alpha_{ij}=\frac{\exp(\boldsymbol{q}_{ij}^{\top}\boldsymbol{k}_{ij}/\sqrt{d})} {\sum_{r\in\mathcal{N}_i}\exp(\boldsymbol{q}_{ir}^{\top}\boldsymbol{k}_{ir}/\sqrt{d})}, \quad \boldsymbol{c}_i=\sum_{j\in\mathcal{N}_i}\alpha_{ij}\boldsymbol{l}_{ij} \label{eq:attention} \end{equation}
The policy is $\boldsymbol{\pi}(\boldsymbol{a}_{i,t}\mid\boldsymbol{o}_{i,t},\boldsymbol{c}_i)$, where $\boldsymbol{o}_{i,t}$ comprises the ego state, neighbor states within sensing radius $d_{\mathrm{in}}$, and previous neighbor actions. Obstacle perception uses a sector-based LiDAR representation (Fig.~\ref{fig:figS2_action_and_obstacle_perception}).

To align local learning with the global pairwise objective, we use a decoupled pairwise energy
\begin{equation}
  \hat{H}_p=\sum_{i\in\mathcal{V}}\frac{1}{|\mathcal{N}_i|}
  \sum_{j\in\mathcal{N}_i}\psi_{ij},
  \label{eq:decoupledPairwise}
\end{equation}
which yields the same triangular-lattice equilibrium as the global pairwise energy $H_p$ (Theorem~\ref{thm:geometry}). This correspondence allows $r_i$ to be computed from local information while remaining aligned with the global objective. From the mean-field perspective, the weighted aggregation in Eq.~\eqref{eq:attention} provides the optimal mean-field approximation (Lemma~\ref{lemma:mean_field}), connecting the local energy-based reward to the decentralized policy without access to the full swarm state.

The system is trained with decentralized training and decentralized execution (DTDE) using proximal policy optimization (PPO), with local policy and value networks. Network hyperparameters and training configurations are given in the Supplementary Materials.

To integrate specialized behaviors without retraining, we use an integration module (Fig.~\ref{fig:figS5_moe_architecture}). Specialist policies are pretrained for traversal, racing, formation, and tracking under task-specific reward potentials (Table~\ref{tab:expert_design}), then frozen. The integration module is implemented as a lightweight gating network that maps local context and a one-hot task command to weights $G_e(\boldsymbol{o},\boldsymbol{c})$, producing
\begin{equation}
  \boldsymbol{\pi}_{\mathrm{U}}(\boldsymbol{o},\boldsymbol{c})
  =\sum_{e=1}^{E}G_e(\boldsymbol{o},\boldsymbol{c})\,\boldsymbol{\pi}_e(\boldsymbol{o}).
  \label{eq:moe}
\end{equation}
The gate is trained with PPO augmented by an exponentially annealed task-instruction loss, so it initially follows the high-level command and progressively relies on local context. Actions are sampled from $\boldsymbol{\pi}_{\mathrm{U}}$ during training and selected by maximum probability at inference. We note that the gating mechanism is one of several possible implementations of policy composition within the energy representation; the contribution of this work lies in the shared energy formulation rather than in the specific integration architecture.


\clearpage 
\bibliography{references}
\bibliographystyle{sciencemag}

\end{document}